\documentclass[12pt]{article}

\usepackage{times}
\usepackage{latexsym}
\usepackage{epsfig}

\usepackage{graphicx}
\usepackage{pstricks, pst-node, pst-tree}
\usepackage{amsmath}

\usepackage{comment}
 
\newcommand{\qed}{\mbox{$\Box$}} 

\newenvironment{INproof}{\begin{trivlist} 
\item[]{\it Proof.\ }}{ %
\mbox{}\hfill\qed 
\end{trivlist}}

\newtheorem{theorem}{Theorem}[section] 
\newtheorem{corollary}[theorem]{Corollary} 
\newtheorem{definition}[theorem]{Definition} 
 
\newtheorem{lemma}[theorem]{Lemma} 
\newtheorem{example}[theorem]{Example}

\newcommand{\Not}{{\sf not} \,}

\long\def\comment#1{}

\newcommand{\inm}{\mbox{in}}
\newcommand{\ta}{\mbox{ta}}
\newcommand{\pa}{\mbox{pa}}
\newcommand{\taol}{\mbox{taol}}
\newcommand{\paol}{\mbox{paol}}
\newcommand{\myalign}{&\hspace{-.4in}&}

\title{\Large Recycling Computed Answers in Rewrite Systems for Abduction\footnote{An extended abstract of parts of this paper appeared in the 
proceedings of IJCAI-03, Acapulco, Mexico.}}

\normalsize
\author{ \large
Fangzhen Lin\\
Department of Computer Science\\
Hong Kong University of Science and Technology\\
Clear Water Bay, Kowloon, Hong Kong \\
\\
Jia-Huai You\\
Department of Computing Science\\
University of Alberta\\
Edmonton, Alberta, Canada T6G 2E8\\
}

\begin{document} 
\bibliographystyle{abbrv} 
\date{}

\maketitle

\normalsize
\begin{abstract}
In rule-based systems, goal-oriented computations 
correspond naturally to the possible 
ways that an observation may be explained.
In some applications, we need
to compute explanations for a series of observations with the same domain.
The question whether previously computed answers can be recycled arises. 
A yes answer could result in substantial savings of repeated computations.
For systems based on classic logic, the answer is {\em yes}.
For nonmonotonic systems however, one tends to believe 
that the answer should be {\em no}, since recycling is a form of adding information.
In this paper, we show that computed answers can always
be recycled, in a nontrivial way, 
for the class of rewrite procedures proposed earlier in
\cite{linyou02} for logic programs with negation.
We present some experimental results on an encoding of the logistics domain.
\end{abstract}

\section{Introduction}
The question we shall address in this paper is the following. With 
a sound 
and complete procedure for abduction, 
suppose we have computed explanations (conveniently represented 
as a disjunction)
$Es = E_1 \vee ... \vee E_n$
for observation $q$. 
Suppose also that in the course of computing explanations for another observation $p$,
we run into $q$ again. Now, we may use the proofs $Es$ for $q$
without actually proving $q$ again.
The question is this: will the use of the proofs $Es$ for $q$ 
in the proof for $p$ preserve the soundness and completeness of the procedure? 

In this paper, we answer this question positively, but in a nontrivial way, 
for the class of rewrite procedures 
proposed in \cite{linyou02} for abduction in logic programming under (partial) stable model semantics (\cite{g-l-88-long}, \cite{prz-90b}).
The main result is a theorem (Theorem~\ref{main-theorem}) that 
says recycling preserves the soundness and completeness.

The general idea of recycling is not new.
Recycling in systems 
based on classic logic is always possible, since 
inferences in these systems can be viewed as 
transforming a logic theory to a logically equivalent one.
In dynamic programming, it is the use of the answers 
for previously computed subgoals that reduces the computational complexity.
In some game playing programs, 
for example in the world champion checker 
program {\em Shinook} (www.cs.ualberta.ca/\verb#~#chinook),
the endgame database stores
the computed results for endgame situations which can be referenced 
in real-time efficiently.

However, the problem of recycling in a nonmonotonic proof system has rarely
been investigated. We note that recycling 
is to use previous proofs.
This differs from adding consequences. 
For example, it is known that the semantics based on
answer sets or (maximal) partial stable models \cite{dun-91}
do not possess the 
{\em cautious nonmonotonicity} property. 
That is, 
adding a consequence of a program
could gain additional models thus losing some consequences.
The following example is due to Dix \cite{dix-91}: 
\begin{equation}
\label{intro}
P = \{a \leftarrow \Not b.  ~~~b \leftarrow c, \Not a. ~~~c \leftarrow a.\}
\end{equation}
$P$ has only one answer set,
$\{a,c\}$. Thus, $c$ is a consequence. When augmented with the rule
$c \leftarrow$, the program gains a second answer set, 
$\{b,c\}$, and loses $a$ as a consequence. 

Abduction in the framework of
logic programming with answer sets \cite{g-l-88-long} 
or partial stable models \cite{prz-90b} has been studied extensively, and
a number of formalisms and top-down query answering procedures have been proposed 
\cite{consoleJLP91,dun-91,dun-95-1,ek89-long,k-m-90,k-k-t-95,k-m-90p,k-m-mJLP00,linyou02,s-i-91,satoh-92-long}.

The class of rewrite procedures for abduction proposed in \cite{linyou02}
is based on the idea of {\em abduction as confluent and terminating rewriting}.
These systems are called {\em canonical systems} in the literature of rewrite 
systems \cite{dershowitz90}.
The confluence and termination properties guarantee that rewriting 
terminates at a unique normal form
independent of the order of rewriting.
Thus, each particular strategy of rewriting 
yields a rewrite procedure.

These rewrite procedures can be used to compute explanations 
using a nonground program,
under the condition that  
in each rule
a variable that appears in the body must also appear in the head. 
Under this condition, an observation (a ground goal) 
is always rewritten to another ground goal, so that a rewriting mechanism
desgined for ground programs works just as well.
When the condition 
is not satisfied, one only needs to instantiate those variables that only appear 
in the body of a rule.
For example, domain restricted programs \cite{niemela-amai-long}
can be instantiated only on domain predicates for variables that do not appear in the head. This is a significant departure from the approaches that are based on
ground computation where a function-free program is first instantiated to 
a ground program with which the intended 
models are then computed.

These rewrite procedures can also be used for answer set semantics in the following way.
If a query $q$ is written into $False$, there cannot be any answer set containing $q$.
This is because answer sets for normal programs are 
special cases of partial stable models.
However, if the query is written into $True$, 
to see whether there is an answer set
containing this query, one then only needs to check whether the 
{\em context} generated so far  can be extended to an answer set,  a task that is
normally much easier than finding an answer set from scratch.
There is a special case, however, when the corresponding propositional 
program is finite and so-called {\em odd-loop} free,  
partial stable
models coincide with stable models. Thus the rewrite procedures are also
sound and complete for these programs.

The next section defines logic program semantics.
Section~\ref{review} reviews the rewriting framework.
Then 
in Section~\ref{result} we formulate 
rewrite systems with computed rules and prove that recycling preserves 
soundness and completeness. Section~\ref{abduction} extends this result
to rewrite systems with abduction, and 
Section~\ref{experiments} reports some experimental results.

\section{Logic Program Semantics}
A rule is of the form 
$$a \leftarrow b_1, ..., b_m, \Not c_1, ..., \Not c_n.$$
where $a$, $b_i$ and $c_i$ are atoms of the underlying propositional
language ${\cal L}$. $\Not c_i$ are called {\em default negations}. \
A {\em literal} is an atom $\phi$ or its negation $\neg \phi$.
A {\em (normal) program} is a finite set of rules.

The {\em completion} of a program $P$, denoted $Comp(P)$,
is a set of equivalences:
for each atom $\phi \in {\cal L}$, 
if $\phi$ does not appear as the head of any rule in $P$,
$\phi \leftrightarrow F \in Comp(P)$; 
otherwise, $\phi \leftrightarrow B_1 \vee ... \vee B_n \in Comp(P)$
(with default negations replaced by the corresponding negative literals)
if there are exactly $n$ rules 
$\phi \leftarrow B_i \in P$ with 
$\phi$ as the head.
We write $T$ for $B_i$ if $B_i$ is empty.
 
The rewriting system of \cite{linyou02} 
is sound and complete w.r.t.\
the partial model semantics \cite{prz-90b}.
A simple way to define partial stable models without even 
introducing 3-valued logic is by the so called {\em alternating fixpoints} \cite{y-y-95}. 
Let $P$
be a program and $S$ a set of default negations.
Define a function over sets $S$ of default negations: 
$F_P (S) = \{ \Not a \, |\, P \cup S \not \vdash a\}$. 
The relation $\vdash$ is the standard propositional derivation relation with
each default negation $\Not \phi$ being treated as a named atom $\Not\_\phi$.

A {\em partial stable model} $M$ is defined by 
a fixpoint of the function that applies $F_P$ twice, 
$F_P^2 (S) = S$, while satisfying $S \subseteq F_P (S)$, 
in the following way: 
for any atom $\xi$, $\neg \xi \in M$ if $\Not \xi \in S$,
$\xi \in M$ if $P \cup S \vdash \xi$, and $\xi$ is {\em undefined}
otherwise. 
An {\em answer set} $E$ is defined by a fixpoint $S$ 
such that $F_P (S) = S$ and $E = \{\xi \in {\cal L} ~|~P \cup S \vdash \xi\}$.

\section{Goal Rewrite Systems}
\label{review}
We introduce goal rewrite systems as formulated in \cite{linyou02}. 

A goal rewrite system 
is a rewrite system that consists of three types of rewrite rules:
(1)
Program rules from $Comp(P)$ for literal rewriting; 
(2)
Simplification rules to transform and simplify goals; and
(3)
Loop rules for handling loops. 

A {\em program rule} is a completed definition 
$\phi \leftrightarrow B_1 \vee \ldots \vee B_n \in Comp(P)$ used from 
left to right: 
$\phi$ can be rewritten to $B_1 \vee \ldots \vee B_n$, and 
$\neg \phi$ to $\neg B_1 \wedge \ldots \wedge \neg B_n$.
These are called {\em literal rewriting}.

A {\em goal}, also called a {\em goal formula},
is a formula which may involve $\neg$, $\vee$ and $\wedge$. 
A goal resulted from a literal rewriting from another goal is 
called a {\em derived goal}. Like a formula, a goal may be transformed 
to another goal without changing its semantics. This is carried out
by simplification rules. 

We assume that in all goals 
negation appears only in front of a literal.  
This can be achieved by simple transformations using the following rules:
for any formulas $\Phi$ and $\Psi$,
$$
\begin{array}{ll}
\neg \neg \Phi \rightarrow \Phi\\
\neg (\Phi \vee \Psi) \rightarrow \neg \Phi \wedge \neg \Psi\\ 
\neg (\Phi \wedge \Psi) \rightarrow \neg \Phi \vee \neg \Psi 
\end{array}
$$
 
\subsection{Simplification rules}

The simplification rules constitute a nondeterministic transformation system
formulated with a mechanism 
of loop handling in mind, which
requires keeping track of 
literal sequences \( g_0, \ldots, g_n \) where
each $g_i$, $0 < i \le n$, is in the goal formula resulted from 
rewriting $g_{i-1}$. Two central mechanisms in
formalizing goal rewrite systems are {\em rewrite chains} and {\em contexts}.

\begin{itemize}
\item
\underline{\em Rewrite Chain:}
Suppose a literal $l$ is written by its definition 
$\phi \leftrightarrow \Phi$
where $l = \phi$ or $l = \neg \phi$. Then,
each literal $l'$ in the derived goal is generated in order to 
prove $l$. This ancestor-descendant relation is denoted 
$l \prec l'$.
A sequence $l_1 \prec \ldots \prec l_n$ is then 
called a {\em rewrite chain}, abbreviated as $l_1 \prec^+ l_n$.

\item
\underline{\em Context:}
A rewrite chain $g = g_0 \prec g_1 \prec \ldots \prec g_n = T$ 
records a set of literals $C = \{g_0, ..., g_{n-1}\}$
for proving $g$.
We will write 
$T(\{g_0, ..., g_{n-1}\})$ and call $C$ a {\em context}.
A context will also be used to maintain consistency:
if $g$ can be proved via a conjunction, 
all of the conjuncts need be proved with 
contexts that are non-conflicting with each other.
For simplicity, we assume that whenever $\neg F$ 
is generated, it is automatically
replaced by $T(C)$, where $C$ is the set of literals on the corresponding 
rewrite chain,
and $\neg T$ is automatically replaced by $F$.
\end{itemize}

Note that for any literal in a derived goal, the 
rewrite chain leading to it from a literal in the given goal 
is uniquely determined.  
As an example, suppose the completion of a program has the definitions: 
$a \leftrightarrow \neg b \wedge \neg c$
and $b \leftrightarrow q \vee \neg p$. Then, we get a rewrite sequence,
$$a \rightarrow \neg b \wedge \neg c \rightarrow \neg q \wedge p \wedge \neg c.$$
For the three literals in the last goal, we have rewrite chains from $a$:
$a \prec \neg b \prec \neg q$; $a \prec \neg b \prec p$; and 
$a \prec \neg c$.

\vspace{.1in}
\noindent
{\bf Simplification Rules:} Let $\Phi$ and $\Phi_i$ be goal formulas, $C$ be a context, and 
$l$ a literal.

\begin{itemize}
\item [SR1.]
$F \vee \Phi \rightarrow \Phi$
\item [SR1']
$\Phi \vee F \rightarrow \Phi$ 
\item [SR2.]
$F \wedge \Phi \rightarrow F$
\item [SR2']
$\Phi \wedge F \rightarrow F$
\item [SR3.]
$T(C_1) \wedge T(C_2) \rightarrow T(C_1 \cup C_2)$~~~ if $C_1 \cup C_2$ is consistent
\item [SR4.]
$T(C_1) \wedge T(C_2) \rightarrow F~~~$ if $C_1 \cup C_2$ is inconsistent
\item [SR5.]
$\Phi_1 \wedge (\Phi_2 \vee \Phi_3) \rightarrow (\Phi_1 \wedge \Phi_2) \vee (\Phi_1 \wedge \Phi_3)$
\item [SR5'.]
$(\Phi_1 \vee \Phi_2) \wedge \Phi_3 \rightarrow (\Phi_1 \wedge \Phi_3) \vee (\Phi_2 \wedge \Phi_3)$ \mbox{}\hfill\qed 
\end{itemize}

SR3 merges two contexts if they contain no complementary literals, otherwise
SR4 makes it a failure to prove. SR4 can be implemented more efficiently by 
$$T(C) \wedge l \rightarrow F~~~\mbox{if}~~\neg l \in C$$
Repeated applications of SR5 and SR5' 
can transform any goal formula to a disjunctive normal form (DNF).

\subsection{Loop rules}
After a literal $l$ is rewritten, it is possible that at some later stage
either $l$ or $\neg l$ appears again in a goal on the same rewrite chain. 
Two rewrite rules are formulated to handle loops.

\begin{definition}
Let $S = l_1 \prec^+ l_n$ be a rewrite chain.

\begin{itemize}
\item
If $\neg l_1 = l_n$ or $l_1 = \neg l_n$, then $S$ is called an {\em odd loop}.
\item
If $l_1 = l_n$, then
\begin{itemize}
\item
$S$ is called {\em a positive loop} if $l_1$ and $l_n$ are both atoms and 
each literal on $\l_1 \prec^+ l_n$ is also an atom;
\item
$S$ is called {\em a negative loop} if $l_1$ and $l_n$ are both negative literals and 
each literal on $\l_1 \prec^+ l_n$ is also negative;
\item
Otherwise, $S$ is called an {\em even loop}. 
\end{itemize}
\end{itemize}
In all the cases above, $l_n$ is called a {\em loop literal}.
\end{definition}

\vspace{.1in}
\noindent
{\bf Loop Rules:}  Let $g_1 \prec^+ g_n$ be a rewrite chain.
\begin{itemize}
\item [LR1.]
$g_n \rightarrow F$\\
if $g_i \prec^+ g_n$, for some $1 \le i < n$,
is a positive loop or an odd loop.
\item [LR2.]
$g_n \rightarrow T(\{g_1, ..., g_n\})$\\
if $g_i \prec^+ g_n$, for some $1 \le i < n$,
is a negative loop or an even loop. 
\mbox{}\hfill\qed 
\end{itemize}

A {\em rewrite sequence} is a sequence of zero or 
more rewrite steps $Q_0 \rightarrow \ldots \rightarrow Q_{k}$,
denoted $Q_0 \rightarrow^* Q_k$, such that 
$Q_0$ is an initial goal,
and for each $0 \le i < k$, 
$Q_{i+1}$ is obtained from $Q_i$ by 
\begin{itemize}
\item
literal rewriting at a non-loop literal in $Q_i$,
or 
\item
applying a simplification rule to a subformula of $Q_i$, or
\item
applying a loop rule to a loop literal in $Q_i$.
\end{itemize}

\vspace{.1in}
\begin{example}
For the program given in the Introduction, 
$$P_0 = \{a \leftarrow \Not b.  ~~b \leftarrow c, \Not a. ~~c \leftarrow a.\}$$
$a$ is proved but $b$ is not. This is shown by the following rewrite sequences:
$$
\begin{array}{ll}
a \rightarrow \neg b \rightarrow \neg c \vee a \rightarrow \neg a \vee a 
\rightarrow F \vee a
\rightarrow a \rightarrow T(\{a, \neg b\})\\
b \rightarrow c \wedge \neg a \rightarrow a \wedge \neg a 
\rightarrow \neg b \wedge \neg a 
\rightarrow F \wedge \neg a \rightarrow F
\end{array}
$$
Let $P_1 = \{b \leftarrow \Not c.\;~ c \leftarrow c.\}$. 
$b$ is proved and $\neg b$ is not.
$$
\begin{array}{ll}
b \rightarrow \neg c \rightarrow \neg c \rightarrow T(\{\neg b,\neg c\});~~~~~
\neg b \rightarrow c \rightarrow c \rightarrow F 
\end{array}
$$
\mbox{}\hfill\qed 
\end{example}

Note that, in general, the proof-theoretic meaning of a goal formula may not be the same as
the logical meaning of the formula.
For example, the goal formula 
$a \vee \neg a$ (a tautology in classic logic)
could well lead to an $F$ if neither $a$ nor $\neg a$ can be proved, e.g.,
for the program $\{ a \leftarrow \Not a\}$.

\begin{definition}
A {\em goal rewrite system} for a program $P$ is a triple 
$\langle {\cal Q}_L, {\cal R}_P, \rightarrow\rangle$,
where ${\cal Q}_L$ is the set of all goals,
${\cal R}_P$ is a set of rewrite rules which 
consists of program rules from $Comp(P)$, 
the simplification rules and the loop rules, and   
$\rightarrow$ is the set of all rewrite sequences. 
\end{definition}

\subsection{Previous results}

Goal rewrite systems are like term rewriting systems \cite{dershowitz90}
everywhere except 
at terminating steps: a terminating step at a subgoal may 
depend on the history of rewriting. 

A set of rewrite sequences defines a binary relation, say $R$,
on the set of goal formulas: $R(Q,Q')$ iff $Q \rightarrow^* Q'$.
Hence, a set of rewrite sequences corresponds to a 
binary relation. 

Two desirable properties of rewrite systems are the properties of 
termination and confluence.
Rewrite systems that possess both of these properties 
are called canonical systems. A canonical system
guarantees that 
the final result of rewriting from any given goal is unique, 
independent of any order of rewriting. 

\begin{definition}
A goal rewrite system $\langle {\cal Q}_L, {\cal R}_P, \rightarrow\rangle$
is {\em terminating} iff there exists no
endless rewrite sequence $Q_1 \rightarrow Q_2 \rightarrow Q_3 \rightarrow ......$ in $\rightarrow$.
\end{definition}

\begin{definition}
A goal rewrite system $\langle {\cal Q}_L, {\cal R}_P, \rightarrow \rangle$
is {\em confluent} iff
for any rewrite sequences 
$t_1 \rightarrow^* t_2$ and $t_1 \rightarrow^* t_3$, 
there exist
$t_4 \in {\cal Q}_L$ and rewrite sequences 
$t_2 \rightarrow^* t_4$ and $t_3 \rightarrow^* t_4$.
\end{definition}

In \cite{linyou02}, it is shown that
all goal rewrite systems defined above are canonical,
i.e., they are confluent and terminating.
It was also shown any goal rewrite 
system is sound and complete w.r.t.\ the partial stable model semantics:

\begin{theorem}  
\label{sound-and-complete}
Let $P$ be a finite program and 
$\langle {\cal Q}_L, {\cal R}_P, \rightarrow \rangle$
a goal rewrite system.

\vspace{.1in}
\noindent
{\sf Soundness:} For any literal $g$ and any rewrite sequence
$g \rightarrow^* T(C_1) \vee \ldots \vee T(C_m)$, 
there exists a partial stable model
$M_i$ of $P$, for each $i \in [1..m]$,
such that $g \in C_i \subseteq M_i$.

\vspace{.1in}
\noindent
{\sf Completeness:} For any literal $g$ true in 
a partial stable model $M$ of $P$,
there exists a rewrite sequence
$g \rightarrow^* T(C_1) \vee \ldots \vee T(C_m)$
such that there exists $i \in [1..m]$, $g \in C_i \subseteq M$.
\label{sound-complete}
\end{theorem}

\section{Goal Rewrite Systems with Computed Rules}
\label{result}
We first use two examples to illustrate the main technical results of this 
paper.

\begin{example}
Given a rewrite system $R^0$, suppose we have a rewrite sequence 
$~\neg q \rightarrow a \rightarrow a \rightarrow F$.
The failure is due to a positive loop on $a$. 
We may recycle
the computed answer by replacing the rewrite rule for $\neg q$ by 
the new rule, $\neg q \rightarrow F$. We thus get a new system, say 
$R^1$. Suppose in trying to prove $g$ we have
$$g \rightarrow a \rightarrow \neg q \rightarrow F$$
where the last step makes use of the computed answer for $\neg q$.
The question arises as whether this way of using previously computed results
guarantees the soundness and completeness.
Theorem~\ref{main-theorem} to be proved later in this paper answers this 
question positively.
To see it for this example,
assume we have the following, successful proof
in $R^0$
$$g \rightarrow a \rightarrow \neg q \rightarrow a \rightarrow T(\{g,a,\neg q\})$$
where the termination is due the even loop on $a$.
\comment{$a \prec \neg q \prec a$.}
Had such a sequence existed, recycling would have produced a wrong result.
However, one can see that the existence of the rewrite step 
$a \rightarrow \neg q$ implies the existence of a different way to
prove $\neg q$:  
$$\neg q \rightarrow a \rightarrow \neg q \vee \ldots \rightarrow T(\{\neg q, a\}) \vee \ldots$$
contradicting that $\neg q$ was rewritten to $F$ in $R^0$. \mbox{}\hfill\qed 
\end{example}

Before giving the next example, we introduce 
a different way to understand rewrite sequences. Since
any goal formula 
can always be transformed to a DNF using the distributive rules SR5 and 
SR5',
and the order of 
rewriting does not matter, we can view 
rewriting as generating 
a sequence of DNFs. Thus,
a rewrite sequence in DNF from 
an initial goal $g$,  
$$g \rightarrow^* N_1 \vee ... \vee N_n$$
can be 
conveniently represented by {\em derivation trees},
or {\em d-trees}, one for each $N_i$ representing one possible way
of proving $g$.
For any $i$, the d-tree for $N_i$
has $g$ as its root node, wherein a branch from $g$ to a leaf node 
corresponds to a rewrite chain 
from $g$ that eventually ends with an $F$ or some $T(C)$. 
As such a disjunct is a conjunction, a successful proof requires 
each branch to succeed
and the union of all resulting contexts to be consistent. 

The next example is carefully constructed to illustrate that 
recycling may not yield the same answers as if no recycling were carried out. In particular, one can sometimes get additional answers.

\begin{example}
\label{main-example}
Consider the program:
\begin{eqnarray*}
\myalign g \leftarrow a.\ \ \ a \leftarrow \Not b.\ \ \ a \leftarrow e.\\ 
\myalign b \leftarrow b.\ \ \ e \leftarrow p. \ \ \ \ \ p \leftarrow a. \\
\end{eqnarray*}
\begin{figure}
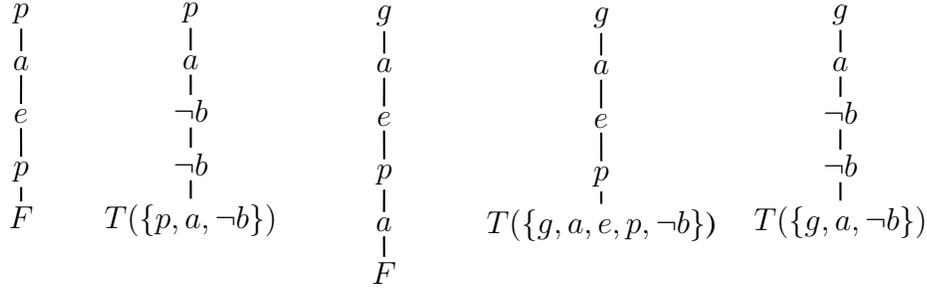

\hspace{.7in}            
\pstree[nodesep=2pt,levelsep=20pt] {\TR{$p$}}
          {\pstree{\TR{$a$}} 
             {\pstree{\TR{$e$}} 
                {\pstree{\TR{$p$}} 
                   {\pstree{\TR{$F$}} 
                  {}}
                {}}
             {}}
}

\vspace{-1.42in}
\hspace{1.2in}
\pstree[nodesep=2pt,levelsep=20pt] {\TR{$p$}}
          {\pstree{\TR{$a$}} 
             {\pstree{\TR{$\neg b$}} 
                {\pstree{\TR{$\neg b$}} 
                   {\pstree{\TR{$T(\{p,a,\neg b\})$}} 
                   {}}
                {}}
             {}}
}

\vspace{-1.4in}
\hspace{2.6in}
\pstree[nodesep=2pt,levelsep=20pt] {\TR{$g$}}
          {\pstree{\TR{$a$}} 
             {\pstree{\TR{$e $}} 
                {\pstree{\TR{$p$}} 
                    {\pstree{\TR{$a$}} 
                        {\pstree{\TR{$F$}} 
                        {}}
                    {}}
                {}}
             {}}
}

\vspace{-1.7in}
\hspace{4.6in} 
\pstree[nodesep=2pt,levelsep=20pt] {\TR{$g$}}
          {\pstree{\TR{$a$}} 
             {\pstree{\TR{$\neg b$}} 
                {\pstree{\TR{$\neg b$}} 
                  {\pstree{\TR{$T(\{g,a,\neg b\})$}} 
                  {}}
                {}}
             {}}
}

\vspace{-1.4in}
\hspace{3.2in}
\pstree[nodesep=2pt,levelsep=20pt] {\TR{$g$}}
          {\pstree{\TR{$a$}} 
             {\pstree{\TR{$e $}} 
                {\pstree{\TR{$p$}} 
                    {\pstree{\TR{$T(\{g,a,e,p,\neg b\}$)} }
                    {}}
                {}}
             {}}
}
\caption{Recycling may generate extra proofs}
\label{cross-over}
\end{figure}
In Fig.~\ref{cross-over}, each d-tree consists of a single branch.
The left two d-trees are expanded from goal $p$
corresponding to the following rewrite sequence:
$$
\begin{array}{ll}
p \rightarrow a \rightarrow e \vee \neg b \rightarrow p \vee \neg b \rightarrow F \vee \neg b
\rightarrow \neg b \rightarrow \neg b \rightarrow T(\{p,a,\neg b\})
\end{array}
$$
The next two d-trees are for goal $g$,
corresponding to the rewrite sequence:
$$
\begin{array}{ll}
g \rightarrow a\rightarrow e \vee \neg b \rightarrow p \vee \neg b \rightarrow a \vee \neg b \\
\rightarrow F \vee \neg b \rightarrow \neg b \rightarrow \neg b \rightarrow T(\{g,a,\neg b\})
\end{array}
$$
Now, we recycle the proof for $p$ in the proof for $g$ 
and compare it with the one without recycling.
Clearly, the successful d-tree for $g$ (the fourth from the left) will still succeed as it doesn't involve any $p$. 
The focus is then on the d-tree in the middle,
in particular, the node $p$ in it; this d-tree 
fails when no recycling was
performed.

Since $p$ is previously proved with context $\{g,a,\neg b\}$, 
recycling of this proof amounts to terminating $p$ with a context which is the union
of this context with the rewrite chain leading to $p$ (see the d-tree on the right).
But this results in a successful proof that fails without recycling. 

Though recycling appears to have generated a wrong result,
one can verify that both generated contexts,
$\{g,a,\neg b\}$ and $\{g,a,e,p,\neg b\}$,
belong to the same 
partial stable model. 
Thus, recycling
in this example didn't lead to an incorrect answer but generated
a redundant one.
Theorem~\ref{main-theorem}
shows that this is not incidental. 
Indeed, if $p$ is true in a partial stable model, by derivation (look at the d-tree
in the middle), so must be $e$, $a$, and $g$.  \mbox{}\hfill\qed 
\end{example}

\subsection{Rewrite systems with computed rules}

Given a goal rewrite system $R$, 
we may denote a rewrite sequence from
a literal $g$ by $g \rightarrow_R E$.
\begin{definition} {\rm (Computed rule)}\\
Let $R$ be a goal rewrite system in which literal $p$ is rewritten to
its normal form. The {\em computed rule for $p$} is defined as:
If $p \rightarrow_R F$, the computed rule for $p$ 
is the rewrite 
rule $p \rightarrow F$; 
if $p \rightarrow_R T(C_1) \vee ...\vee T(C_n)$,
then the computed rule for $p$ is the rewrite 
rule $p \rightarrow T(C_1) \vee ...\vee T(C_{n})$.
\end{definition}

For the purpose of recycling, a computed rule $p \rightarrow E$ 
is meant to replace the existing literal rewrite rule for $p$. 
If a computed rule $p \rightarrow F$ representing a failed derivation,
it can be 
used directly as the literal rewrite rule for $p$.
Otherwise, we must combine the contexts in $E$ with the rewrite chain leading
to $p$, and keep only consistent ones. 

\vspace{.1in}
\noindent
\underline{\bf Recycling Rule:}  \\
Let $g_1 \prec^+ g_n$ be a rewrite chain where $g_n$ is a non-loop literal.
Let $G = \{g_1,...,g_n\}$, and 
$g_n \rightarrow T(D_1) \vee ... \vee T(D_k)$ be the computed rule for 
$g_n$. Further, let $\{D'_1,...,D'_{k'}\}$ be the subset 
of $\{D_1,...,D_k\}$ containing any $D_i$ such that 
$D_i \cup G$ is consistent.
Then, the {\em recycling rule} for $g_n$
is defined as:

\begin{itemize}
\item [RC.]
$g_n \rightarrow T(G \cup D'_1) \vee ... \vee T(G \cup D'_{k'})$ \mbox{}\hfill\qed 
\end{itemize}

\begin{example}
Consider the following program:
\begin{eqnarray*}
\myalign g \leftarrow a. ~~~~~~~~~~a \leftarrow p.\ \ \ ~~~~~~~p \leftarrow \Not a.\\ 
\myalign a \leftarrow \Not p.\ \ ~~ p \leftarrow \Not b. \ \ \ \ \ b \leftarrow \Not a. 
\end{eqnarray*}
and the proof:
$$
\begin{array}{ll}
p \rightarrow \neg a \vee \neg b \rightarrow p \vee \neg b \rightarrow
T(\{p,\neg a\}) \vee \neg b \\
\rightarrow
T(\{p,\neg a\}) \vee p \rightarrow T(\{p,\neg a\}) \vee T(\{p,\neg b\})
\end{array}
$$ 
We therefore have a computed rule for $p$:
$$p \rightarrow T(\{p,\neg a\}) \vee T(\{p,\neg b\})$$
Now, in the course of proving $g$ we can recycle the computed rule for $p$:
$$
\begin{array}{ll}
g \rightarrow a \rightarrow p \vee \neg p \rightarrow T(\{g,a, p,\neg b\}) \vee p 
\rightarrow ...
\end{array}
$$
\end{example}

In the sequel, a rewrite system includes 
the recycling rule as well as zero or more 
computed rules. We note that 
the termination and confluence properties remain to hold
for the extended systems.

We are interested in the soundness and completeness of 
a series of rewrite systems, each of which  
recycles computed answers generated on the previous one.
For this purpose, given a program $P$ we use $R_P^0$ to denote the 
original goal rewrite system where literal rewrite rules
are defined by the Clark completion of $P$. 
For all $i \ge 0$,
$R_P^{i+1}$ is defined in terms of $R_P^i$ as follows:
Let $\Delta_i$ be the set of computed rules 
(generated) on $R_P^i$ for the set of literals ${\cal L}_{\Delta_i}$
Then, $R_P^{i+1}$ is the rewrite system obtained from 
$R_P^i$ by replacing the rewrite rules for the literals in ${\cal L}_{\Delta_i}$ by those in $\Delta_i$. In the rest of this sectin,
we will always refer to a fixed program $P$. Thus we may drop the subscript $P$ and write $R^i$.

\begin{definition}
A rewrite system $R^i$ is {\em sound} iff,  
for any literal $g$ and rewrite sequence 
$g \rightarrow_{R^i} T(C_1) \vee ... \vee T(C_n)$, 
and for each $C_j$, $j \in [1..n]$, 
there exists a partial 
stable model $M$ of $P$ such that $g \in C_j \subseteq M$.
$R^i$ is {\em complete} iff, for any literal $g$ such that
$g \in M$ for some partial stable model $M$ of $P$, there is a rewrite
sequence 
$g \rightarrow_{R^i} T(C_1) \vee ... \vee T(C_n)$ such that for some
$C_j$, $j \in [1..n]$, $g \in C_j \subseteq M$.
\end{definition}

An important property of provability by rewriting 
is the so-called {\em loop rotation}, which is needed in order to prove
the completeness of recycling; namely, a proof (a successful branch 
in a d-tree) terminated by a loop rule can be captured in rotated forms.

To describe this property, we need the following notation about rewrite chains:
Any direct dependency relation $l \prec l'$ may be 
denoted by $l \cdot l'$, and we allow a segment (which may be empty) 
of a rewrite chain to be denoted 
by a Greek letter such as $\delta$, $\theta$, and $\xi$.
Thus, we may write
$x \cdot \delta \cdot y$ to denote a rewrite chain from $x$ to $y$ via $\delta$,
or $x \cdot \delta$ 
to mean a rewrite chain that begins with $x$ followed by the segment 
denoted by $\delta$. 
A rewrite chain may also be used to denote the set of the literals 
on it.

\begin{figure}
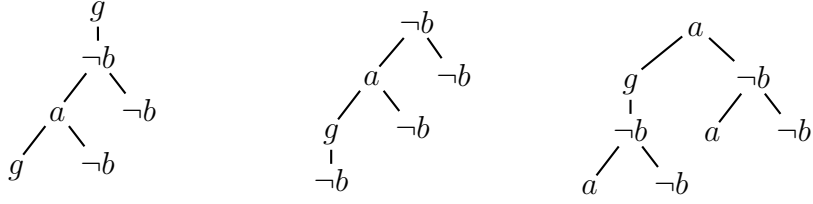

\hspace{1in}            
\pstree[nodesep=2pt,levelsep=20pt] {\TR{$g$}}
  {\pstree{\TR{$\neg b$}} 
          {\pstree{\TR{$a$}} 
               {\TR{$g$}
                \TR{$\neg b$}
               }
           \pstree{\TR{$\neg b$}}  
           {}
          }
  }

\vspace{-0.9in}
\hspace{2.6in}
\pstree[nodesep=2pt,levelsep=20pt] {\TR{$\neg b$}}
  {\pstree{\TR{$a$}} 
     {\pstree{\TR{$g$}} {\TR{$\neg b$}} 
      \pstree{\TR{$\neg b$}} 
      {}
     }
   \TR{$\neg b$} 
  {}
  }

\vspace{-0.9in}
\hspace{4in}
\pstree[nodesep=2pt,levelsep=20pt] {\TR{$a$}}
  {\pstree{\TR{$g$}} 
     {\pstree{\TR{$\neg b$}} 
       {\TR{$a$} \TR{$\neg b$}} 
     }
   \pstree{\TR{$\neg b$}}
       {\TR{$a$} \TR{$\neg b$}} 
  {}
  }

\caption{Loop rotation}
\label{rotation}
\end{figure}

\begin{lemma} (loop rotation)\\
Let $R^0$ be a rewrite system without computed rules.
Let $Tr$ be a d-tree for literal $g$ that succeeds with context $C$.
Suppose a branch of $Tr$ ends with a loop, $g \cdot \theta \cdot g$, for
some $\theta$. 
Then, for any literal $l \in \theta$, there is a proof 
of $l$ that succeeds with the same context $C$.
\end{lemma}
\begin{INproof}
A loop, $\pi = g \cdot l_1 \cdot l_2 \cdot \ldots \cdot  l_n \cdot g$, 
where $g$ and $l_i$ are literals,
can always be rotated as 
$$
\begin{array}{ll}
l_1 \cdot l_2 \cdot \ldots \cdot l_n \cdot g \cdot l_1,\ \ \ \ \ \\
l_2 \cdot \ldots \cdot l_n \cdot g \cdot l_1 \cdot l_2,\ \ \ \ \\......
\end{array}
$$
and so on,
so that if $\pi$ is a negative loop (or an even loop, resp.) so is its
rotated loop. Rotation over a d-tree can be performed 
as follows: remove the top node $n$,
and for any link from the top node, $n \cdot q$, attach the link $n \cdot q$ 
to any occurrence of $n$. The assumption of the existence of 
loop $g \cdot \theta \cdot g$ ensures that 
in every round of rotation there is 
at least one occurrence of the top node. (See Fig~\ref{rotation} 
for an illustration where rotation proceeds from left to right.)
It can be seen that the type of a loop is always preserved and the set of 
literals on the tree remains unchanged. 
\label{rotation-lemma}
\end{INproof}

\subsection{Soundness and completeness of recycling}

Below, given a literal $l$, by {\em a proof of $l$} we mean a
rewrite sequence from $l$ to 
$T(C_1) \vee \ldots \vee T(C_n)$,
where any $C_i$ can be referred to as a proof of $l$.

\begin{theorem}
\label{main-theorem}
For any $i \ge 0$, $R^i$ is sound and complete.
\end{theorem}

\begin{INproof} 
We prove the claim by induction on $i$. 
$R^0$, the system without computed rules, 
is sound and complete \cite{linyou02}.
Now assume for all $j$ with $0 \le j \le i$, $R^j$ are sound and complete, and show that 
$R^{i+1}$ is also sound and complete.

We only need to consider the situations where rewriting in $R^{i+1}$ 
differs from that of $R^i$. 
Let ${\cal L}_{\Delta_i}$ be the set of literals whose computed rules are generated on $R^i$.
We can first carry out rewriting
without rewriting the literals that are 
in ${\cal L}_{\Delta_i}$. In this case, 
rewriting from $g$ in both $R^i$ and $R^{i+1}$ 
terminate at the same expression, which is either $F$ or a DNF, say
$N_1 \vee... \vee N_m$. Each $N_i$ can be represented
by a d-tree.

\vspace{.1in}
\noindent
{\em Soundness:} 
Suppose $g \rightarrow_{R^{i+1}} T(D_1) \vee ... \vee T(D_s)$. For any 
$D \in \{D_1,...,D_s\}$ we need to show that there is a partial stable 
model $M$ such that $D \subseteq M$.
Consider the d-tree $Tr$ that generates $D$ and suppose $g$ is
its root node.
We show inductively in a bottom-up fashion that all the literals on 
$Tr$ must be in the same partial stable model. 
For any leaf node $p$
that is terminated by its computed rule 
$$p \rightarrow ... \vee T(C) \vee...$$
suppose $Tr$ is the one
that succeeded with context $C$.
By the inductive hypothesis on $R^j$,
we know that $R^j$ is sound for all $j \le i$,
thus there is a partial stable model $M$ such that $C \subseteq D \subseteq M$.
If a leaf node $q$ is terminated by a loop, by the loop rotation lemma
(lemma~\ref{rotation-lemma}), there is a proof of $q$ in $R^i$ 
using rotated loops. Otherwise we have an obvious case where
a leaf node is rewritten to $True$ by its Clark completion.

In the inductive step, let $l_1,...,l_n$ be the child nodes of some node $l$ 
and assume each $l_i$ is proved in $R^i$ hence in some partial stable model. 
We first show that they belong to the same partial stable model $M$. Then,
we show that $l$ can also be proved in $R^i$ thus belonging to $M$ 
as well.
Without loss of generality, assume
there are only two child nodes:
$l_1 \rightarrow_{R^i} T(Q_1) \vee ... \vee T(Q_m)$, 
$l_2 \rightarrow_{R^i} T(W_1) \vee ... \vee T(W_n)$.
Since $D$ is constructed in $R^{i+1}$ using computed rules, by definitions
of computed rule and the recycling rule,
there are $Q_i$ and $W_j$ such that
$Q_i \cup W_j \subseteq D$, and hence $Q_i \cup W_j$ is consistent.
Then in $R^i$, the two contexts are merged by using simplification rule SR3, i.e.,
$$l_1 \wedge l_2 \rightarrow_{R^i} ... \vee [T(Q_i) \wedge T(W_j)] \vee ...
\rightarrow_{R^i} ... \vee T(Q_i \cup W_j) \vee ...$$
Since $R^i$ is sound, there is a partial stable model $M$
such that $\{l_1,l_2\} \subseteq Q_i \cup W_j \subseteq M$.
But $l$ is derivable from $l_1$ and $l_2$. Using the definition of 
partial stable models, it can be shown 
that $l$ must also be in $M$.

The induction allows us to conclude that 
for the top goal $g$ and its proof $D$ in $R^{i+1}$, we must have 
$g \in D \subseteq M$, for the same partial stable model $M$.

\vspace{.1in}
\noindent
{\em Completeness:} 
We show that for any context generated in $R^i$, 
the same context will be generated in $R^{i+1}$. Then, 
$R^{i+1}$ is complete simply because $R^i$ is complete.

Let $p \in {\cal L}_{\Delta_i}$, and consider a proof of $g$ via $p$ and its
d-tree.
Since each branch of this d-tree can be expanded and eventually terminated 
independent of others, for simplicity, we consider a proof of $g$ simply by 
(an extension of) a branch $g \cdot \xi \cdot p$.  
In $R^{i+1}$ the computed rule for $p$ is used while 
in $R^i$ it is not. 
We only need to consider two cases of proof in 
$R^i$: 
either $g$ is proved via $p$ and a previously computed rule, 
or the proof is terminated due to a loop.

\vspace{.1in}
\noindent
(i) The case of loops. In expanding the rewrite chain 
$g \cdot \xi \cdot p$ in $R^i$,
we may form a loop, say $g \cdot \xi \cdot p \cdot \xi'$. If the loop is in $\xi'$,
exactly the same loop occurs in rewriting $p$ as the top goal in $R^i$,
so it is part of the computed rule for $p$. Otherwise it is a
loop that {\em crosses over p}, in the general form
$$\pi = g \cdot \theta_1 \cdot l \cdot \theta_2 \cdot p \cdot \theta_3 \cdot l$$
where $l$ is the loop literal. As a special case of loop rotation over a branch
(cf. Lemma~\ref{rotation-lemma}),
the same way of 
terminating a rewrite chain presents itself in proving $p$ 
as the top goal in $R^i$, which is
$$\pi' = p \cdot \theta_3 \cdot l \cdot \theta_2 \cdot p.$$ If the loop on $\pi$ is a
negative loop (or an even loop, resp.), so is $\pi'$. 
Thus the same context will be generated in $R^{i+1}$.

\vspace{.1in}
\noindent
(ii) $g$ is proved via $p$ and a previously computed rule. That is, $R^i$
gives a rewrite chain of the form 
$g \cdot \xi \cdot p \cdot \delta \cdot q$ where $q \rightarrow E$ is a computed rule generated on $R^j$ for some $j < i$. 
Suppose the context generated this way
is $C$.
Because of the existence of $p \cdot \delta \cdot q$, exactly the same computed rule 
$q \rightarrow E$ must be used in generating the computed rule for $p$ in $R^i$.
It can be seen that the context generated in $R^{i+1}$
by recycling the computed answers for $p$ (which is computed via $q$)
is exactly the same as the one that uses the computed answers for
$q$ but not those for $p$.
So, for any context generated this way in $R^i$,
the same context will be generated in $R^{i+1}$ as well.
\end{INproof}

As given in the corollary below, if we only recycle failed proofs then exactly the same contexts will
be generated.
\begin{corollary}
\label{fail_case}
Let $R^i$ be a rewrite system where each computed rule is of the 
form $p \rightarrow F$. Let $g$ be a literal and $E$ be a normal form.
Then, for any $i \ge 0$,
$g \rightarrow_{R^0} E~~~\mbox{iff}~~~g \rightarrow_{R^i} E$.
\end{corollary}

\begin{INproof}
Let $\Delta$ be the set of literals whose rewrite rules are 
computed rules in $R^i$. Consider rewriting without rewriting on the literals
in $\Delta$. Then, rewriting from $g$ terminates at the 
same expression $E'$, which is either an $F$ or $T(C_1) \vee ... \vee T(C_n)$,
in both $R^0$ and $R^i$. The claim then follows
from the theorem above that for any $q \in \Delta$, $q \rightarrow_{R^0} F$
iff $q \rightarrow_{R^i} F$. That is, if $q \rightarrow_{R^0} F$, 
then $q$ is not in any partial stable model. 
The soundness of $R^i$ ensures that if $q \rightarrow_{R^i} Q$ where $Q \not = F$, then there is a partial stable model containing $q$, resulting in
a contradiction. The converse is similar.
\end{INproof}

\section{Recycling in Abductive Rewrite Systems}
\label{abduction}
As shown in \cite{linyou02}, the rewriting framework
can be extended to 
abduction in a straightforward way: the only difference in the extended 
framework is that we do not apply the Clark
completion to abducibles. 
That is,
once an abducible appears in
a goal, it will remain there unless it is eliminated by
the simplification rule $SR2$ or $SR2'$. In a similar way, the goal rewrite
systems with computed rules in the previous section can be extended to
abduction as well.

\begin{definition} {\rm (Computed rule for abduction)}\\
Let $R$ be an extended goal rewrite system for abduction.
The {\em computed rule for $p$} is defined as:
If $p \rightarrow_R F$, the computed rule for $p$ 
is the rewrite 
rule $p \rightarrow F$; 
if 
\begin{eqnarray*}
&& p \rightarrow_R [l_{11}(C_{11})\land\cdots\land l_{1k_{1}}(C_{1k_{1}})] \vee \ldots \vee\\
&&\hspace*{2em}  [l_{m1}(C_{m1})\land\cdots\land l_{mk_{m}}(C_{mk_{m}})]
\end{eqnarray*}
such that each $l_{ij}$ is either $T$ or 
an abducible literal, and
$C_{i1}\cup\cdots\cup C_{ik_i}$ is consistent for each $i$,
then the computed rule for $p$ is the rewrite 
rule 
\begin{eqnarray}
&& p \rightarrow [l_{11}(C_{11})\land\cdots\land l_{1k_{1}}(C_{1k_{1}})] \vee \ldots \vee \nonumber \\
&&\hspace*{2em}  [l_{m1}(C_{m1})\land\cdots\land l_{mk_{m}}(C_{mk_{m}})]
\label{computedrule}
\end{eqnarray}
\end{definition}
\vspace{.1in}
\noindent
\underline{\bf Recycling Rule:}  \\
Let $g_1 \prec^+ g_n\prec p$ be a rewrite chain where $p$ 
is a non-loop literal.
Let $G = \{g_1,...,g_n,p\}$, and (\ref{computedrule})
be the computed rule for 
$p$. 
Then, the {\em recycling rule} for $p$
is defined as:
\begin{itemize}
\item [RC'.]
\begin{eqnarray*}
&& p \rightarrow [l_{11}(C_{11}\cup G)\land\cdots\land l_{1k_{1}}(C_{1k_{1}}\cup G)] \vee \ldots \vee  \\
&&\hspace*{2em}  [l_{m1}(C_{m1}\cup G)\land\cdots\land l_{mk_{m}}(C_{mk_{m}}\cup G)]
\end{eqnarray*}
\end{itemize}

\section{A Recycling Strategy}

We have shown that in theory, one can reuse the previously
computed answers in our rewrite systems for abduction. To put the
theory into practice, we need some effective strategies on how
to recycle these computations. 

If we want to compute the abduction of all goals in a set, without the
framework of recycling introduced here, the only way is to compute them
one by one independently. With the idea of recycling, we can try to 
recycle previously computed answers. The question is then which goals
to compute first. This question arises even if we just want to
compute the abduction of a single goal:
instead of computing it using the original program, it may sometimes
be better if we first
compute the abduction of some other goals and recycle the results.

Assuming that goals are literals, a
simple strategy for deciding the order of goals to be computed
is to find out the dependency relations among the goals.

\begin{definition}
A literal $l$ is said to be depending on a literal $l'$ if the atom in
$l$ depends on the atom in $l'$. 
An atom $p$ is said to be depending on an atom
$q$ if either $q$ is in
the body of a rule whose head is $p$ or inductively, there is another
atom $r$ such that $p$ depends on $r$, and $q$ is in the body
of a rule whose head is $r$.
\end{definition}

It is easy to see that if $l$ depends on $l'$, but $l'$ does not depend
on $l$, then $l'$ will never be sub-goaled to $l$ during rewriting, but
$l$ could be sub-goaled to $l'$. Thus if we need to compute the abduction
of both $l$ and $l'$, we should do it for $l'$ first.

\section{Experiments}
\label{experiments}
We have implemented a depth-first search rewrite procedure with 
branch and bound. The procedure can be used to 
compute explanations using a nonground program,
under the condition that  
in each rule
a variable that appears in the body must also appear in the head. 
When this condition 
is not satisfied, one only needs to instantiate those variables that only appear 
in the body of a rule.
This is a significant departure from the approaches that are based on
ground computation where a function-free program is first instantiated to 
a ground program with which the intended 
models are then computed.

To check the
effectiveness of the idea of recycling, we tested our system on the
logistics problem in \cite{linyou02}.
This is a domain in which there is a truck and a package. A package can be in
or outside a truck, and a truck can be moved from one location to another.
The problem is that given state constraints such as
that the truck and the package can each be at only one location at any given
time, and that if the package is in the truck, then when the truck moves
to a new location, so does the package, how we can derive
a complete specification of the effects of the action of moving a truck
from one location to another. Suppose that we have the following 
propositions:
$\ta(x)$ ($\pa(x)$) \--- the truck (package) is at location $x$ initially;
$\inm$ \--- the package is in the truck initially;
$\ta(x,y,z)$ ($\pa(x,y,z)$)
\--- the truck (package) is at location $x$ after performing
the action of moving
it from $y$ to $z$;
$\inm(y,z)$ \--- the package is in the truck after performing
the action of moving
the truck from $y$ to $z$.
Then in \cite{linyou02},
the problem is solved by computing the abduction of successor state
propositions
$\{ \ta(x,y,z), \pa(x,y,z), \inm(y,z)\}$ in terms of initial state propositions
$\{ \ta(x), \pa(x), \inm\}$ (abducibles)
using the following logic program:
\begin{eqnarray}
\myalign \ta(X,X1,X). \label{ta1}\\
\myalign
\pa(X,X1,X2) \leftarrow \ta(X,X1,X2), \inm(X1,X2). \label{pa1}\\
\myalign \ta(X,X1,X2) \leftarrow X \neq X2,
    \ta(X), \Not \taol(X,X1,X2). \label{ta2}\\
\myalign
\taol(X,X1,X2) \leftarrow Y \neq X, \ta(Y,X1,X2).\\
\myalign \pa(X,X1,X2) \leftarrow 
    \pa(X), \Not \paol(X,X1,X2). \label{pa2}\\
\myalign \paol(X,X1,X2) \leftarrow  Y \neq X, \pa(Y,X1,X2).\\
\myalign \inm(X,Y) \leftarrow \inm. \label{in}
\end{eqnarray}
Here the variables are to be instantiated over a domain of locations.
For instance, given query $\pa(3,2,3)$, our system would compute its
abduction as  $\pa(3)\lor\inm$, meaning that for 
it to be true, either
the package was initially at $3$ or it was inside the truck.

According to the definition in the last section,
literals that contain
$\pa(X,Y,Z)$ depend on those that contain
$\inm(X,Y)$ and  $\ta(X,Y,Z)$.
But literals that contain
$\inm(X,Y)$ and those that contain $\ta(X,Y,Z)$ do not depend on 
each other.
So we should compute first the abduction of $\inm(X,Y)$ and $\ta(X,Y,Z)$.
Now $\inm(X,Y)$ is solved by rule
(\ref{in}), $\ta(X,Y,X)$ by rule (\ref{ta1}), and as it turned out,
when $X\neq Z$, $\ta(X,Y,Z)$ is always false, and its computation is 
relatively easy.
 For instance,
for the domain with 9 locations, query $\ta(7,1,6)$ took only 2.6 seconds.
In comparison, query $\pa(7,1,7)$ took more than 7000 seconds without
recycling.

Table~\ref{data} contains run time data for some representative 
queries.\footnote{Our implementation was written in Sicstus Prolog, and
the experiments were done on a PIII 1GHz notebook with 512 MB memory. For
generating explanations for regular rewriting system, our implementation
is a significant improvement over the one in \cite{linyou02}. For instance, 
for a domain with 7 locations query $\pa(3,2,3)$ took more than
20 minutes for
the implementation reported in \cite{linyou02}, but required less than
1 second under our implementation running on a comparable machine.}
For comparison purpose, 
each query is given two entries:
the one under ``NR'' refers to regular rewriting system 
without using recycling,
and the one under ``WR'' refers to rewriting system using computed
rules about $\ta(X,Y,Z)$. As one can see, especially for hard queries like
$\pa(7,1,7)$, recycling in this case significantly speeds up the 
computation.
\begin{table}
\begin{tabular}{|c|c|c||c|c|}
\hline Query & \multicolumn{2}{|c||}{9 locations} &
\multicolumn{2}{|c|}{10 locations} \\
\cline{2-5} & NR & WR & NR & WR\\
\hline
pa(1,2,3)  &     0.71 &   0.41 &     1.50 &  0.89 \\
-pa(1,2,3) &    75.89 & 2.28 &  342.96 & 5.65    \\
pa(3,2,3)  &    137.05 &  0.89 &  630.69 & 1.98         \\
-pa(3,2,3) &      2.97 & 1.98 &   7.64 &  5.03           \\
pa(1,5,7) &  122.87 &0.75 & 278.07 &1.31    \\
-pa(1,5,7) & 727.6 & 7.07 &2534.09 & 19.08      \\
pa(7,5,1) &  108.66 &  17.82 & 188.50 & 30.72        \\
-pa(7,5,1) &  74.43 &2.26 & 340.51 &5.64          \\
pa(7,1,7) & 7619.72 &  20.78 & 29140.69 & 35.65       \\
-pa(7,1,7) &  2.98 &2.01 &   7.71 &5.05       \\
\hline
\end{tabular}
\caption{Recycling in logistics domain. Legends: NR - no recycling; WR - 
recycling $\ta(X,Y,Z)$ goals. All times are in CPU seconds.}
\label{data}
\end{table}

\section{Concluding remarks and future work}

We have considered the problem of how to reuse previously computed
 results
for answering other queries in the abductive rewriting system of 
Lin and You \cite{linyou02} for logic programs with negation, and showed
that this can indeed be done. We have also described a methodology of 
using the recycling system in practice by analysing the dependency relationship
among propositions in a logic programs. We applied this methodology to
the problem of computing the effect of actions in a logistics domain, the
same one considered in \cite{linyou02}, and 
our experimental results showed that recycling in this domain can 
indeed result in good performance gain.

For future work we are looking for more domains to try our system on and
to implement a system that can
automatically analyse a program and decide how best to recycle previous
computations.

\bibliography{you}
\end{document}